\documentclass[10pt,twocolumn,letterpaper]{article}

\usepackage{iccv}
\usepackage{times}
\usepackage{epsfig}
\usepackage{graphicx}
\usepackage{amsmath}
\usepackage{amssymb}
\usepackage{mathtools}
\usepackage{color}
\usepackage{algorithm}
\usepackage{algpseudocode}
\usepackage{setspace}
\usepackage{multirow}
\usepackage{booktabs}
\usepackage{wrapfig}
\usepackage{subfigure}
\usepackage{kotex}
\usepackage{xcolor}
\usepackage{pifont}
\usepackage{booktabs}
\usepackage{threeparttable}
\usepackage{derivative}
\usepackage{gensymb}
\usepackage{axessibility}
\usepackage[pagebackref=true,breaklinks=true,colorlinks,bookmarks=false]{hyperref}

\newcommand{\hjk}[1]{\textcolor{black}{#1}}

\iccvfinalcopy 

\def\iccvPaperID{11129} 

\ificcvfinal\fi

\newcommand{\omitme}[1]{}

\newcommand{\Rb}{\mathbb{R}}

\newcommand{\pb}{\mathbf{p}}

\newcommand{\argminU}{\mathop{\mathrm{argmin}}}

\def\eg{\emph{e.g}.}
\def\ie{\emph{i.e}.}

\def\Pc{\mathcal{P}}

\newtheorem{theorem}{Theorem}
\newtheorem{proposition}[theorem]{Proposition}

\makeatletter
\newcount\my@repeat@count
\newcommand{\myrepeat}[2]{%
  \begingroup
  \my@repeat@count=\z@
  \@whilenum\my@repeat@count<#1\do{#2\advance\my@repeat@count\@ne}%
  \endgroup
}
\makeatother
\begin{document}

\title{Point Cloud Augmentation with Weighted Local Transformations}

\newcommand*\samethanks[1][\value{footnote}]{\footnotemark[#1]}
\author{Sihyeon Kim${}^{1}$\thanks{Equal contribution. ${}^\dagger$ is the corresponding author.}, Sanghyeok Lee${}^1$\samethanks, Dasol Hwang${}^1$\\
Jaewon Lee${}^1$, Seong Jae Hwang${}^2$, Hyunwoo J. Kim${}^{1\dagger}$\\
{\small ${}^1$Korea University ${}^2$University of Pittsburgh}\\
{\tt\small \{sh\_bs15, cat0626, dd\_sol, 2j1ejyu, hyunwoojkim\}@korea.ac.kr}\\
{\tt\small sjh95@pitt.edu}
}


\maketitle
\ificcvfinal\thispagestyle{empty}\fi

\begin{abstract}
Despite the extensive usage of point clouds in 3D vision, relatively limited data are available for training deep neural networks.
Although data augmentation is a standard approach to compensate for the scarcity of data, it has been less explored in the point cloud literature.
In this paper, we propose a simple and effective augmentation method called PointWOLF for point cloud augmentation.
The proposed method produces smoothly varying non-rigid deformations by locally weighted transformations centered at multiple anchor points.
The smooth deformations allow diverse and realistic augmentations.
Furthermore, in order to minimize the manual efforts to search the optimal hyperparameters for augmentation, we present AugTune, which generates augmented samples of desired difficulties producing targeted confidence scores.
Our experiments show our framework consistently improves the performance for both shape classification and part segmentation tasks.
Particularly, with PointNet++, PointWOLF achieves the state-of-the-art \textbf{89.7} accuracy on shape classification with the real-world ScanObjectNN dataset.
The code is available at https://github.com/mlvlab/PointWOLF.
\end{abstract}

\vspace{-10pt}

\section{Introduction}
\begin{figure}[ht] 
\centering

\includegraphics[trim=0 80 0 0, clip,width=2.2cm]{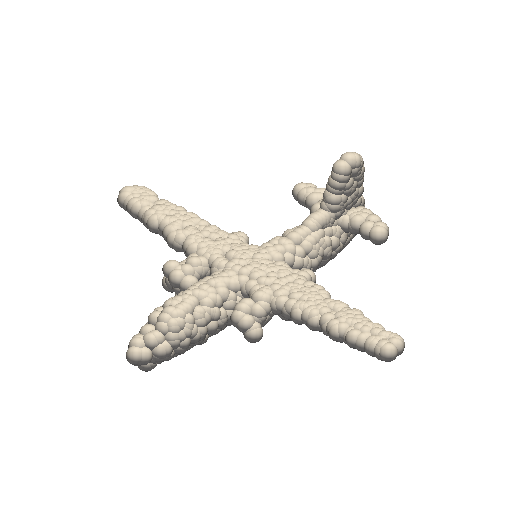}%
\includegraphics[trim=0 80 0 0, clip,width=2.2cm]{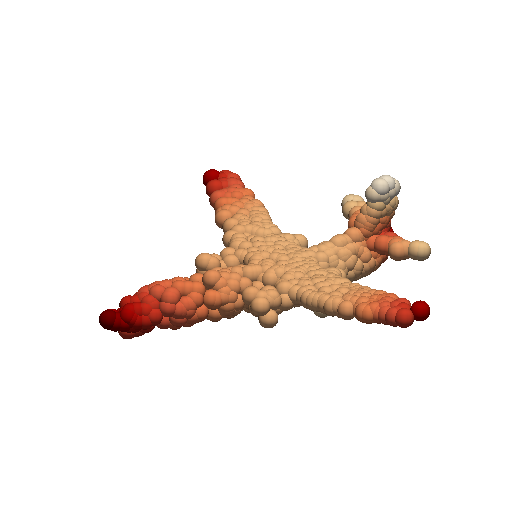}%
\includegraphics[trim=0 80 0 0, clip,width=2.2cm]{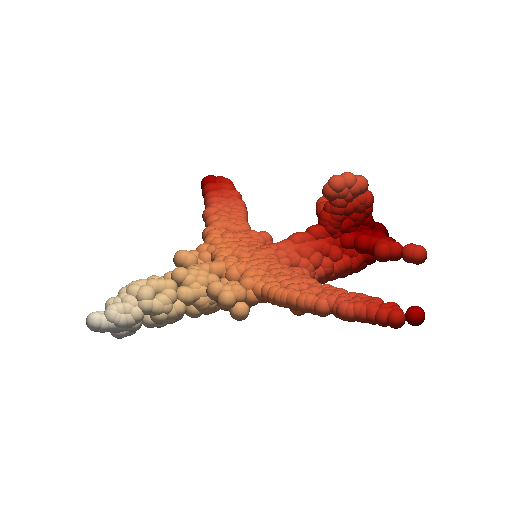}%
\includegraphics[trim=0 80 0 0, clip,width=2.2cm]{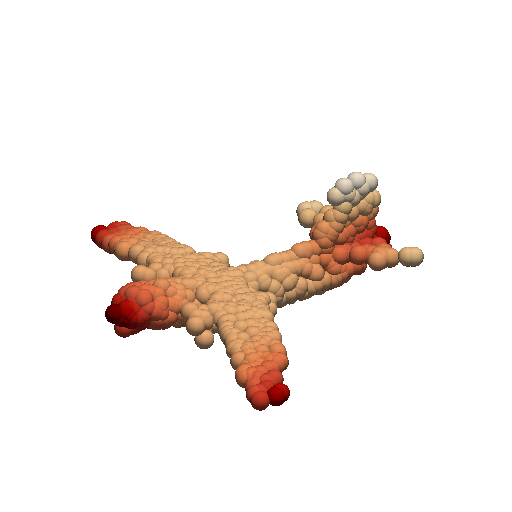}

\includegraphics[trim=-70 30 0 100, clip,width=2.5cm]{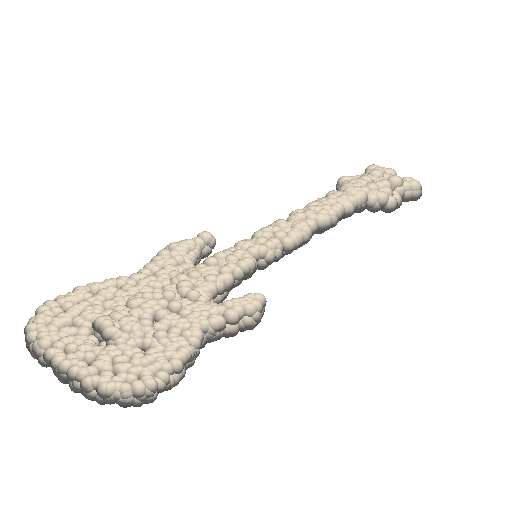}%
\includegraphics[trim=0 30 0 100, clip,width=2.2cm]{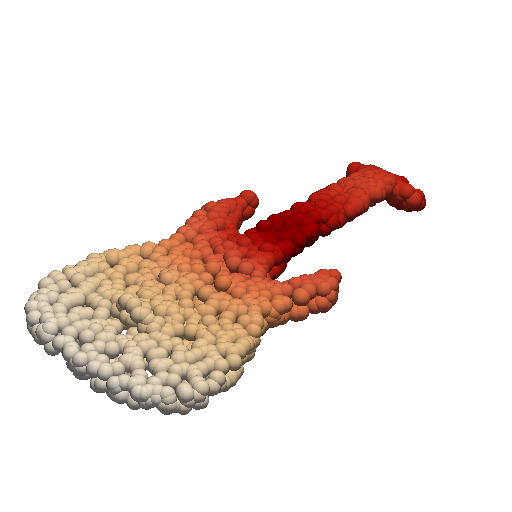}%
\includegraphics[trim=0 30 0 100, clip,width=2.2cm]{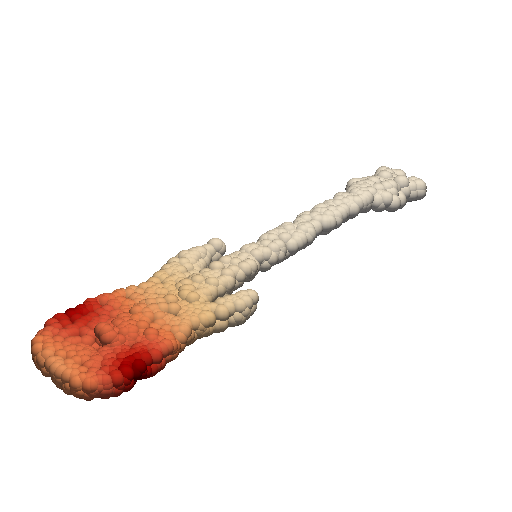}%
\includegraphics[trim=0 30 0 100, clip,width=2.2cm]{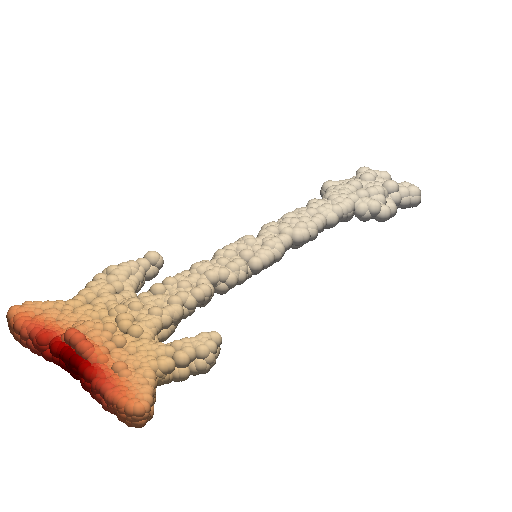}

\includegraphics[height=2.2cm]{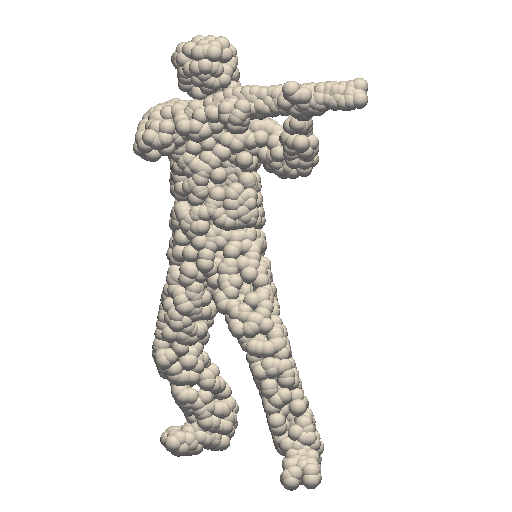}%
\includegraphics[height=2.2cm]{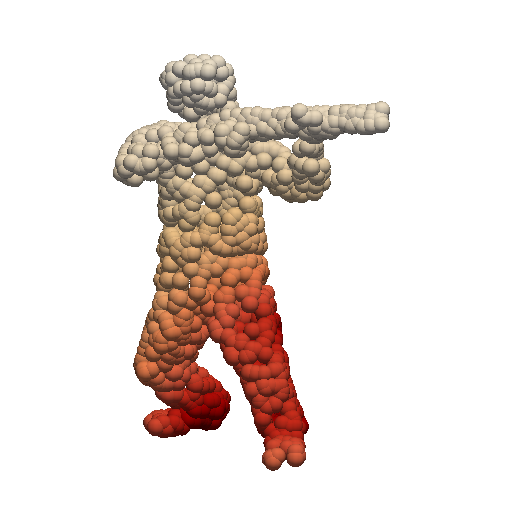}%
\includegraphics[height=2.2cm]{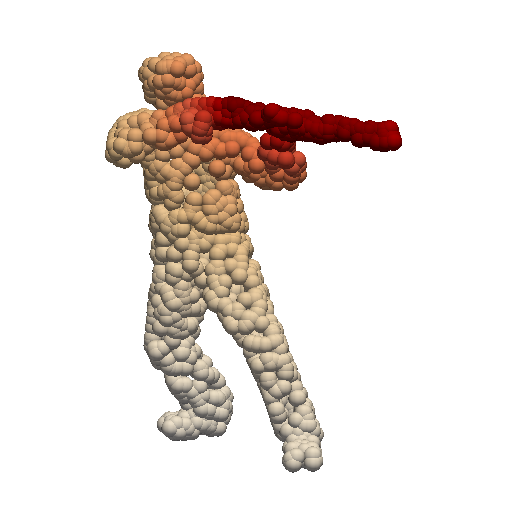}%
\includegraphics[height=2.2cm]{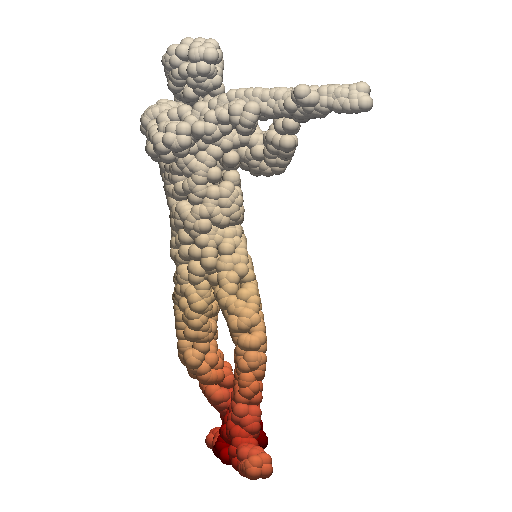}

\includegraphics[height=2.2cm]{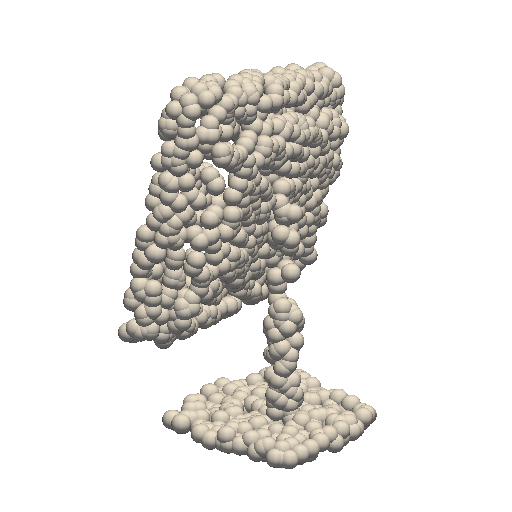}%
\includegraphics[height=2.2cm]{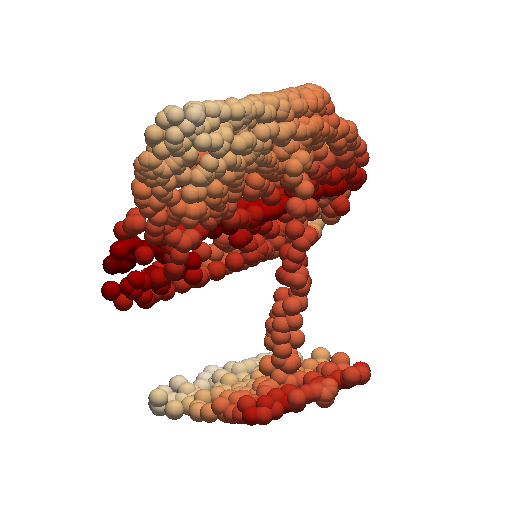}%
\includegraphics[height=2.2cm]{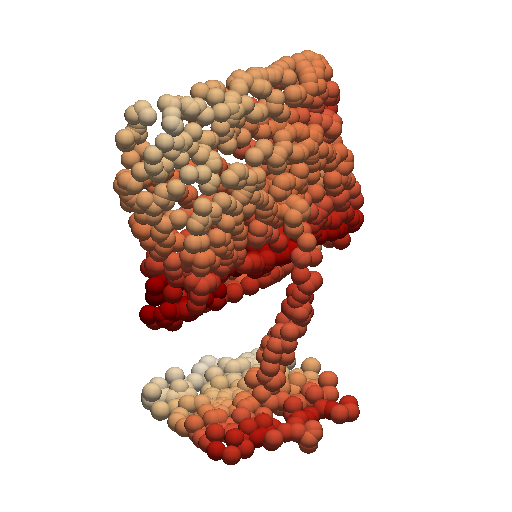}%
\includegraphics[height=2.2cm]{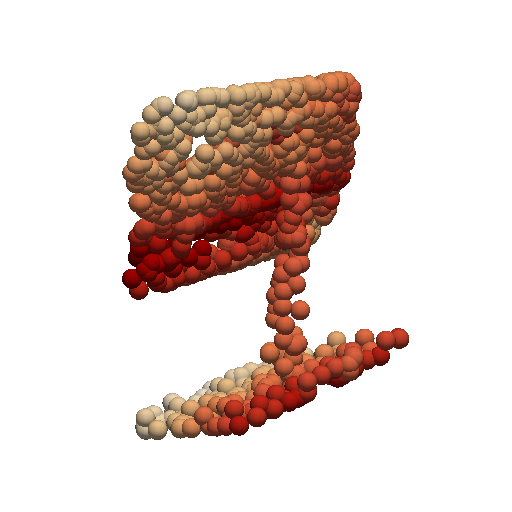}

\includegraphics[height=2.2cm]{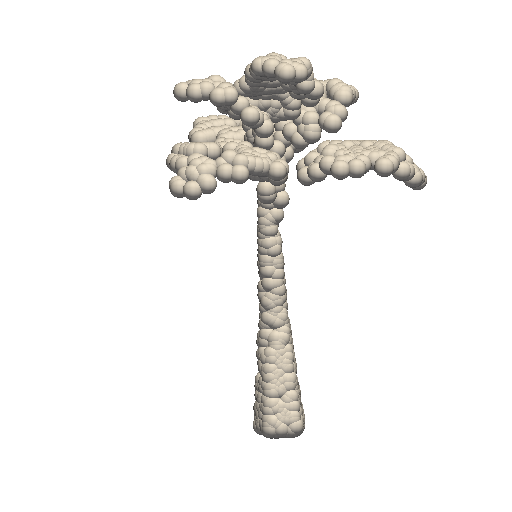}%
\includegraphics[height=2.2cm]{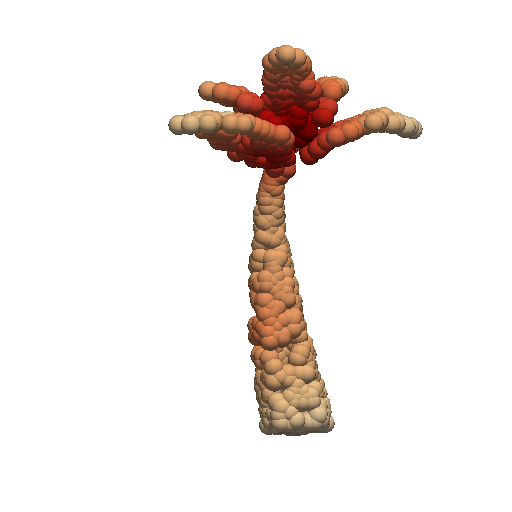}%
\includegraphics[height=2.2cm]{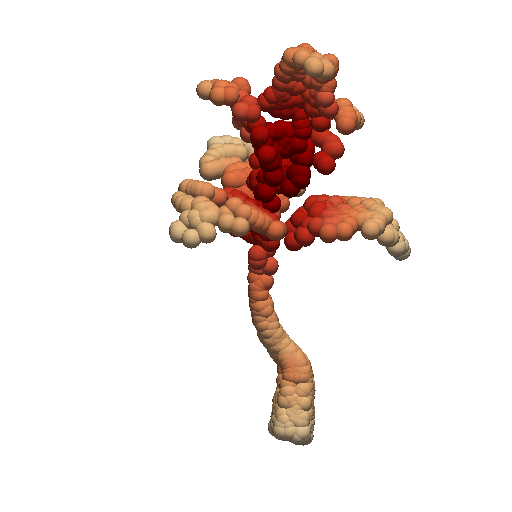}%
\includegraphics[height=2.2cm]{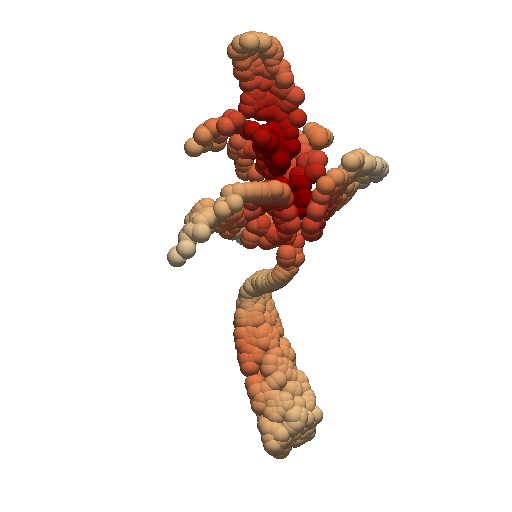}

\vspace{-5pt}
\caption{
\textbf{Locally augmented point clouds using PointWOLF.} In each row, the left most sample is the original, and the remaining samples are its locally transformed results (brighter regions indicate stronger local transformations). PointWOLF can \textit{locally} transform objects while preserving the original shape identity.}
\label{fig:fig1}
\vspace{-15pt}
\end{figure}

Modern deep learning techniques, which established their popularity on structured data, began showing success on point clouds.
Unlike images with clear lattice structures, each point cloud is an unordered set of points with no inherent structures that globally represent various 3D objects.
Recent deep learning efforts have focused on enabling neural networks to operate on point clouds. 
While several point cloud datasets appeared, a particular dataset of scanned real-world objects \cite{scanobjectnn} required a much greater understanding of the point cloud structures to identify highly complex real-world objects.
In response, the approaches have evolved from extracting point-wise information with no structural information \cite{pointnet} to explicitly encoding the \textit{local structure} \cite{pointnet2}. 
These works on network development have been making steady progress despite the scarcity of point cloud data. 

Our interest lies in data augmentation, which is extensively utilized in other machine learning pipelines for solving the data scarcity issue.
Interestingly, despite its prevalence in other data modalities, data augmentation (DA) on point clouds is relatively less explored.
For instance, conventional data augmentation (CDA) \cite{ pointnet, pointnet2}, which consists of global rotation, scaling, translation, and small point-wise noise, is commonly applied to point cloud datasets. 
Recently, PointAugment \cite{pointaugment} proposes to learn the transformation matrix with an augmentor network to produce augmentations. 
PointMixup \cite{pointmixup} finds linearly interpolated point cloud samples and their classes (\eg, Mixup \cite{mixup}).
Despite their efforts to elevate the previous CDA, they still have apparent limitations.
Specifically, PointAugment relies on a single (thus global) transformation matrix/vector for each sample, and PointMixup mixes up the samples globally without explicitly considering each sample's local structures.
Thus, the need for a point cloud augmentation approach capable of producing diverse samples that \textit{accurately depict real-world local variability} (\eg, airplanes with varying lengths of wings and body) still remains.

In this work, we propose a novel point cloud augmentation PointWOLF that satisfies such needs. PointWOLF generates diverse and realistic local deformations such as a person with varying postures (see Figure~\ref{fig:fig1}). 
Our approach systematically enables local deformation by first considering multiple local transformations with respect to anchor points and carefully combining them in smoothly varying manners. 
Furthermore, we present AugTune to adaptively control DA strength in a sample-wise manner. 
AugTune produces consistent and beneficial samples during training with a single hyperparameter which alleviates the known dependence on hyperparameter selection of augmentation. 
We believe our method can further resolve this common dependence issue in general data augmentation.

Our \textbf{contributions} is fourfold:
(\textbf{i}) We propose a powerful point cloud transformation approach capable of generating diverse and realistic augmented samples by deforming local structures.
(\textbf{ii}) Our framework adaptively adjusts the strength of augmentation with only a single hyperparameter.
(\textbf{iii}) We demonstrate that our framework brings consistent improvements over existing state-of-the-art augmentation methods on both synthetic and real-world datasets in point cloud shape classification and part segmentation tasks.
(\textbf{iv}) Our framework improves the robustness of models against various local and global corruptions.

\section{Related Work}
\noindent\textbf{Deep Learning on Point Clouds.}
Early deep learning works on point cloud have focused on enabling existing CNNs to operate on point clouds.
These include multi-view based methods like ~\cite{multiview, volumetic_multiview, multiview_convolution, multiview-hamonized} where they project the 3D point cloud to 2D space through bird's-eye view or multi-view where 2D convolution becomes feasible.
Similarly, other works~\cite{Minkowski, Point_voxel_CNN, FCGF, modelnet, voxelnet} voxelize the point cloud to directly apply 3D convolution on the voxelized point cloud. 
To preserve the original structure of point clouds, point-based methods have emerged. 
PointNet~\cite{pointnet} considers each point cloud as an unordered set and derives point-wise features with multi-layer perceptron and max pooling.
However, a symmetric function such as pooling cannot characterize the local structure of point clouds, thus, PointNet++~\cite{pointnet2} appeared which utilizes local information through hierarchical sampling and grouping. 
Other related studies~\cite{pointwise_CNN, pointCNN, RSCNN, kpconv, DGCNN, pointconv} also rely on grouping to identify the relationship between points and extract local structure.
DGCNN~\cite{DGCNN} explicitly leverages the graph-like structure of the point clouds in the feature space rather than the 3D space.
Interestingly, despite these network-wise efforts to exploit the local structure, only a few works have looked for solutions outside the networks, \eg, data augmentation.

\noindent\textbf{Data Augmentation.} 
Data augmentation (DA) has become a necessity for modern machine learning model training to improve the generalization power.
For point clouds, global similarity transformations such as rotation, scaling, and translation with point-wise jittering~\cite{ pointnet, pointnet2} are conventionally used.
However, such conventional DAs (CDA) do not augment \textit{local} structures which many successful works mentioned above find beneficial and explicitly leverage.
In light of this, only a few recent studies proposed more advanced DAs on point cloud.
PointAugment~\cite{pointaugment} learns an augmentor network to augment samples to be more difficult to classify than the original sample. 
PointMixup~\cite{pointmixup}, enables Mixup~\cite{ mixup,manimix} technique to point cloud, specifically by interpolating between two point cloud samples and predicting the ratio of the mixed classes with a soft label.
While these works enable augmentations beyond simple similarity transformations, the transformations are fundamentally \textit{global}: PointAugment learns a sample-wise global transformations matrix and PointMixup globally interpolates between samples.
Thus, they often do not produce augmentations that are truly local and realistic. In response to this need, we propose a novel augmentation method PointWOLF which \textit{locally} transforms samples as in Figure~\ref{fig:fig1}.

\noindent\textbf{Searching for Optimal DA.}
In practice, identifying strong candidate transformations and optimal parameters for DA lacks intuitive conventions and heuristics thus requires extensive searching process.
Several works address this, for instance, AutoAugment~\cite{autoaugment} and Fast AutoAugment~\cite{Fastautoaugment} dynamically search for the best transformation policy via costly solvers such as reinforcement learning or bayesian optimization.
RandAug~\cite{randaug} has drastically reduced the search space by binding multiple augmentation parameters as a single hyperparameter.
{In this paper, we present AugTune that efficiently controls the sample-wise DA strength with a single parameter using the target confidence score.}

\section{Method}
\begin{figure*}[ht] 
\centering
 \includegraphics[width=1\textwidth]{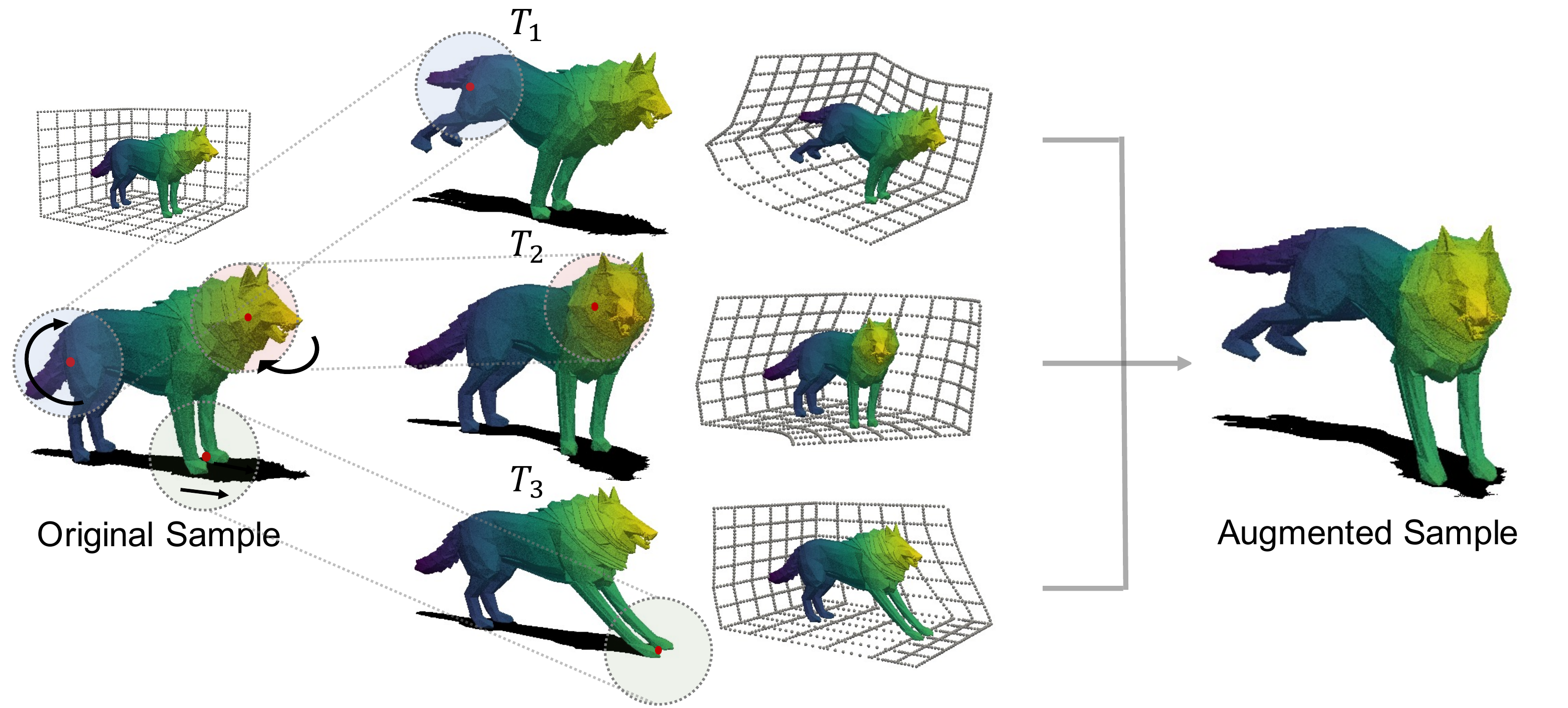}
\vspace{-5pt}
\caption{\textbf{PointWOLF Framework Illustration.} Given an original sample, PointWOLF has multiple local transformations at each anchor point ({\color{red} red}). 
PointWOLF produces smoothly varying non-rigid deformations based on the weighted local transformations.}
\label{fig:fig2}
\vspace{-10pt}
\end{figure*}

We first briefly describe the conventional DA for point clouds.
Then, we describe \textit{PointWOLF}, which aims to generate augmented point clouds. Unlike previous works that perform a global transformation and point-wise jittering, our framework augments the point clouds by locally weighted multiple transformations.
We generate diverse and realistic augmented samples by deforming local structures.
Also, to reduce the dependence on optimal hyperparameters of DA frameworks, 
we present \textit{AugTune}, adaptively modulates the augmentation strength with a single parameter.

\subsection{Preliminaries}
A point cloud $\mathcal{P} \in \mathbb{R}^{(3+C) \times N}$ in 3D is a set of points $\{\pb_1,\cdots,\pb_N\}$. 
Each point is represented as a vector $\pb_i \in \mathbb{R}^{3+C}$ which is a concatenation of its coordinates (\ie, [x,y,z]) and $C$ dimensional input features such as color and a surface normal.
Since the problem of our interest only focuses on the point cloud structure, we assume that no additional input features are given, \ie, $\mathcal{P} \in \Rb^{3\times N}$. 
A conventional data augmentation (CDA) \cite{pointnet,pointnet2} for point clouds applies a {\em global} similarity transformation (\eg, scaling, rotation and translation) and point-wise jittering.
A resulting augmented point cloud ${\mathcal{P}^\prime} \in \Rb^{3\times N}$ is given as follows:
\begin{equation}\small
\mathcal{P}^\prime = s\mathbf{R} \mathcal{P} + \mathbf{B},
\label{eq:cda}
\end{equation}
where $s>0$ is a scaling factor, $\mathbf{R}$ is a 3D rotation matrix, and $\mathbf{B} \in \Rb^{3 \times N}$ is a translation matrix with global translation and point-wise jittering.
Typically, $\mathbf{R}$ is an extrinsic rotation parameterized by a uniformly drawn Euler angle for up-axis orientation. 
Scaling and translation factors are uniformly drawn from an interval, and point-wise jittering vectors are sampled from a truncated Gaussian distribution.

Thus, CDA is simply a similarity transformation with small jittering that cannot simulate diverse shapes and deformable objects.
Unlike synthetic datasets like ModelNet~\cite{modelnet} and ShapeNet~\cite{shapenet}, a real-world dataset like ScanObjectNN~\cite{scanobjectnn} further necessitates the generation of sophisticated deformations such as a mixture of local transformations.
These are exemplified in Figure \ref{fig:fig1}: airplanes with varying lengths and directions of wings and body, guitars in varying sizes and aspect ratios, and people with different heights and postures (e.g., crossing legs).
\subsection{PointWOLF}
We present a simple yet effective \textbf{point} cloud augmentation with \textbf{w}eighted l\textbf{o}ca\textbf{l} trans\textbf{f}ormations (PointWOLF).
Our method generates deformation for point clouds by a convex combination of multiple transformations with smoothly varying weights.
PointWOLF first selects several anchor points and locates random local transformations (\eg, similarity transformations) at the anchor points.
Based on the distance from a point in the input to the anchor points, our method differentially applies the local transformations.
The smoothly varying weights based on the distance to the anchor points allow spatially continuous augmentation and generate realistic samples.
Our framework can be viewed as a kernel regression with transformations.

\noindent\textbf{Sampling anchor points} is the first step of our framework to locate multiple local transformations.
To minimize the redundancy between local transformations, the anchor points $\mathcal{P}^{\mathcal{A}} \subset \mathcal{P}$ are selected by the Farthest Point Sampling (FPS) algorithm.
FPS randomly chooses the first point and then sequentially chooses the farthest points from previous points.
This maximizes the coverage of anchor points and allows diverse transformations.

\noindent\textbf{Local transformations} in our framework are centered at the anchor points.
At each anchor point, we randomly sample a local transformation that includes scaling from the anchor point, changing aspect ratios, translation, and rotation around the anchor point. 
This subsumes the global transformation in \eqref{eq:cda}.
Given an anchor point $\pb^\mathcal{A}_j$ in $\mathcal{P}^{\mathcal{A}}$, the local transformation for an input point $\pb_i$ can be written as:
\begin{equation}\small
    \pb^j_i = \mathbf{S}_j \mathbf{R}_j(\pb_i - \pb^\mathcal{A}_j) + \mathbf{b}_j + \pb^\mathcal{A}_j,
\label{eq:transform_single}
\end{equation}
where $\mathbf{R}_j$, $ \mathbf{S}_j$ and $\mathbf{b}_j$ are rotation matrix, scaling matrix and translation vector $\mathbf{b}_j$ respectively which specifically correspond to $\pb^\mathcal{A}_j$. 
$\mathbf{S}$ is a diagonal matrix with three positive real values, \ie, $\mathbf{S}=\text{diag}(s_x, s_y,s_z)$ to allow different scaling factors for different axes. 
In order to change the aspect ratios along arbitrary directions, a randomly rotated scaling matrix, \eg,  $\mathbf{S}_j'=\mathbf{R}^{-1}\mathbf{S}_j\mathbf{R}$, can be used. 
Since many commonly used datasets are pre-aligned in a standard space (\eg, airplanes facing the same direction), we may assume sensible object orientations. 
In practice, we see that scaling with reasonable rotations as in \eqref{eq:transform_single} is sufficient.

\noindent\textbf{Smooth deformations} are key to generate realistic and locally transformed samples.
A na\"ive application of a random local transformation within its finite neighborhood may result in a discontinuous shape and an overlap of different parts. It has a high chance to lose discriminative structures.  
Instead, we employ the Nadaraya-Watson kernel regression~\cite{nadaraya, watson} to smoothly interpolate the local transformations in the 3D space. 
Given $M$ local transformations $\{T_j\}_{j=1}^M$, our smoothly varying transformation at an arbitrary point $\pb_i$ is given as:
\begin{equation}\small
\hat{T}(\pb_i)= \frac{\sum_{j=1}^{M}{K_{h}}(\pb_i, \pb^\mathcal{A}_j)T_j}  {\sum_{k=1}^{M}{K_{h}}(\pb_i, \pb^\mathcal{A}_k)},
\label{eq:NW}
\end{equation} 
where $K_h(\cdot,\cdot)$ is a kernel function with bandwidth $h$, and $T_j$ is the local transformation in \eqref{eq:transform_single} centered at $\pb^\mathcal{A}_j$. 
To define $\hat{T}(\pb_i)$ at any point in the 3D space,  we use a kernel function that has a strictly positive value for any pair of points, 
\ie, $K_{h}(\pb_i, \pb_j) > 0$ for $\forall \pb_i, \forall \pb_j$.
The following proposition theoretically guarantees that our augmentation is a smooth transformation under mild conditions. The proof is in the supplement.
\begin{proposition} If a kernel function $K_h(\cdot, \cdot)$ and all local transformations $\{T_j\}_{j=1}^M$ are smooth, then the locally weighted transformation $\hat{T}(\cdot)$ in \eqref{eq:NW} is a smooth transformation.
\end{proposition}

In our experiments, we use the Gaussian kernel with Euclidean distance after projection. Our kernel function is 
    \begin{equation}\small
        K_{h}(\pb_i , \pb^\mathcal{A}_j;\Pi_j) = \text{exp}\left(\frac{-\lVert \Pi_j(\pb_i  - \pb^\mathcal{A}_j) \rVert ^{2}_{2}} {2h^2}\right),  
    \label{eq:NW2}
    \end{equation}  
where $h$ is the bandwidth and $\Pi_j \in \Rb^{3 \times 3}$ is a projection matrix. 
$\Pi = \text{diag}(\pi_x, \pi_y, \pi_z)$ is constructed with $\pi_x, \pi_y, \pi_z \sim \text{Bernoulli}(\beta)$ for $\beta \in (0,1)$\footnote{To prevent the projection matrix from zero-matrix, we resample $\Pi$ if (0,0,0) is selected.}, which acts as a ``binary mask'' to measure distances with respect to a random subset of the coordinates. 
For instance, a kernel function with $\Pi_j = \text{diag}(0, 0, 1)$ attenuates the influence of local transformation $T_j$ based on the distance from $\pb^\mathcal{A}_j$ along the $z$-axis, and this allows more diverse and realistic transformations such as shearing and torsion (Section \ref{sec:4.4}) \hjk{by a combination of local similarity transformations.}
Similar to the scaling matrix $\mathbf{S}$ in \eqref{eq:transform_single}, in our experiments we use the projections onto the canonical axes/planes instead of an arbitrary subspace. Our preliminary experiments show that they are sufficient for pre-aligned point clouds.

\setlength{\textfloatsep}{8pt}
\begin{algorithm}[t]\footnotesize
\caption{\label{algorithm_WOLF} \textbf{PointWOLF}} 
\textbf{Input:} original point cloud $\mathcal{P} \in \Rb^{3 \times N}$\newline
\textbf{Input:}  $\#$ points $N$, \; $\#$ anchor points $M$, \;kernel bandwidth $h$ \newline 
\textbf{Input:} range for scaling $\rho_{s}$, range for rotation $\rho_{r}$, range for translation $\rho_{t}$, axis dropout probability $\beta$\newline 
\textbf{Output:} augmented point cloud $\mathcal{P}^{\prime} \in \Rb^{3 \times N}$
\setstretch{1.05}
\begin{algorithmic}[1]
\State $\mathcal{P}^\mathcal{A} \;\leftarrow \; \text{FPS}(\mathcal{P}, \; M)$ \Comment{$\mathcal{P}^\mathcal{A} \in \Rb^{3 \times M}$}
\For {$j=1$ to $M$} 
    \State $\mathbf{S}_j \leftarrow \text{diag}(s_x, s_y, s_z)$ \Comment {$s \sim \text{U}_{[1, \rho_{s}]}$}
    \State $\mathbf{R}_j \leftarrow \text{RotationMatrix}(\theta_x, \theta_y, \theta_z)$ \Comment {$\theta \sim \text{U}_{[-\rho_{r}, \rho_{r}]}$}
    \State $\mathbf{b}_j \leftarrow (b_x, b_y, b_z)$ \Comment {$b \sim \text{U}_{[-\rho_{t}, \rho_{t}]}$}
    \State $\mathbf{\Pi}_j \leftarrow (\pi_x, \pi_y, \pi_z)$ \Comment {$\pi \sim \text{Bernoulli}(\beta)$}
\EndFor
\For {$i=1$ to $N$} 
    \For {$j=1$ to $M$} 
        \State $\pb_{i}^{j} \leftarrow \mathbf{S}_j \mathbf{R}_j (\pb_i-\pb^\mathcal{A}_j) + \mathbf{b}_j + \pb^\mathcal{A}_j$  \Comment{Eq. \eqref{eq:transform_single}}
        \State $w_{i}^{j} \leftarrow K_{h}(\pb_i , \pb^\mathcal{A}_j;\Pi_j) $ \Comment {Eq. \eqref{eq:NW2}}
    \EndFor
    \State $\pb^{\prime}_i \leftarrow \frac{\sum_{j=1}^{M}w^j_i \; {\pb}_{i}^{j} }{\sum_{k=1}^{M} w^k_i}$ \Comment{Eq. \eqref{eq:NW}}
\EndFor
\State $ \mathcal{P}^{\prime} \leftarrow \{\pb^{\prime}_i\}^N_{i=1}$
\end{algorithmic} 
\end{algorithm}

We have introduced our framework from a \textit{kernel regression} perspective.
Figure~\ref{fig:fig2} shows a pipeline of our framework where the Augmented Sample is obtained by  combining local transformations as a smooth transformation $\hat{T}$ and applying it on the Original Sample.
Interestingly, at a high-level, we may also view our framework as an adaptive \textit{interpolation of multiple globally transformed point clouds} resulting from applying different (local) transformations (\eg, $T_1$, $T_2$, $T_3$ in Figure~\ref{fig:fig2}) on the Original Sample.
Thus, our framework can be implemented in two ways: \textbf{(1)} Transforming  each point once by a smoothly varying transformation $\hat{T}$ in Eq.~\eqref{eq:NW} and \textbf{(2)} Transforming each point $M$ times by the local transformations $\{T_j\}_{j=1}^M$ and interpolate these $M$ augmented points by the adaptive weights $K(\pb, \pb^\mathcal{A}_j) / \sum_k K(\pb, \pb^\mathcal{A}_k)$.
Although both approaches require $O(MN)$ complexity if we mainly consider $M$ anchor points and $N$ points, the second approach is slightly more efficient in practice since this only involves operations on points (vector) while the first approach involves operations on transformation matrices.
Thus, we show the pseudocode of the second approach in Algorithm \ref{algorithm_WOLF} and show the first approach's pseudocode in the supplement.

\subsection{AugTune: Effective DA Tuning Method}
The keys to effective data augmentation are strong candidate transformations and the optimal strength of the augmentation.
We introduced PointWOLF that generates more diverse and smooth nonlinear transformations. 
Now, we present an efficient scheme to adaptively adjust the strength of data augmentation during training with a \textbf{single} hyperparameter.
We believe that our scheme benefits not only our framework but also any classical data augmentation methods that heavily rely on an exhaustive grid search with a huge number of hyperparameters.
\vspace{7pt}
\begin{table*}[t]
  \centering 
  \caption{\textbf{Overall accuracy on ModelNet40.}}
  \label{MN40} 
\setlength{\tabcolsep}{10.5pt}
\renewcommand{\arraystretch}{1.05}
  \begin{tabular}{cl|cccc|c}
    \toprule
Dataset & Model & CDA & CDA (w/o R) & PointAugment~\cite{pointaugment} & PointMixup~\cite{pointmixup} & \textbf{PointWOLF}\\
    \midrule
    \multirow{3}*{MN40} 
    & PointNet & 89.2 & 89.7 & 90.8 & 89.9 & \textbf{91.1} \\
    & PointNet++ & 91.3 & 92.5 & 92.4 & 92.7 & \textbf{93.2} \\
    & DGCNN  & 91.7 & 92.7 & 92.9 & 93.1 & \textbf{93.2} \\
    \midrule
    \multirow{3}*{\shortstack{ReducedMN40}} 
    & PointNet   & 81.9   & 82.7   & 84.1  &  83.4  &  \textbf{85.7}\\
    & PointNet++ & 85.9 & 87.8 & 87.0  &  88.6   & \textbf{88.7} \\
    & DGCNN      & 87.5   & 88.8   & 88.3 &  89.0   & \textbf{89.3} \\
    \bottomrule
  \end{tabular}
\end{table*} 
    
\begin{table*}[t]
  \centering 
  \caption{\textbf{Overall accuracy on ScanObjectNN.}}
  \label{SONN}
\setlength{\tabcolsep}{18pt}
\renewcommand{\arraystretch}{1.05}
\begin{tabular}{cl|ccc|c}
    \toprule
    Dataset & Model & CDA & PointAugment~\cite{pointaugment} & PointMixup~\cite{pointmixup} & \textbf{PointWOLF} \\
    \midrule
     \multirow{3}*{\shortstack{OBJ\_ONLY}}  
    & PointNet         & 76.1 & 74.4 & -  & \textbf{78.7} \\
    & PointNet++ & 86.6 & 85.4 & 88.5 & \textbf{89.7} \\
    & DGCNN            & 85.7 & 83.1 & -    & \textbf{88.8}\\
    \midrule
    \multirow{3}*{\shortstack{PB\_T50\_RS}}
    & PointNet         & 64.0 & 57.0 & -    & \textbf{67.1} \\
    & PointNet++ & 79.4 & 77.9 & 80.6 & \textbf{84.1} \\
    & DGCNN            & 77.3 & 76.8 & -  &\textbf{81.6} \\
    \bottomrule
  \end{tabular}

\end{table*}

\begin{algorithm}[t]\footnotesize
\caption{{\label{algorithm_augtune}} \textbf{AugTune}} 
\textbf{Input:} original point cloud $\mathcal{P} \in \Rb^{3 \times N}$, ground truth $\mathbf{y}$ \newline
\textbf{Input:} classifier $f(\cdot;\mathbf{w})$, difficulty coefficient $\lambda \in (0,1]$
\newline
\textbf{Output:} Final augmented point cloud $\mathcal{P}^{\ast}  \in \Rb^{3 \times N}$
\setstretch{1.05}
\begin{algorithmic}[1]
\State $\mathcal{P}^{\prime}  \; \leftarrow \ \text{PointWOLF}(\mathcal{P})$ \Comment{Algorithm \ref{algorithm_WOLF}}
\State  $\mathbf{\hat{y}}^{}_\mathcal{P} \leftarrow \; f(\mathcal{P};\mathbf{w}), \; \mathbf{\hat{y}}^{}_{\mathcal{P}^{\prime}} \leftarrow \; f(\mathcal{P}^{\prime};\mathbf{w})$ 
\State $c_\mathcal{P}^{} \leftarrow {\mathbf{\hat{y}}_\mathcal{P}}^\top  \mathbf{y}$, \; $c_{\mathcal{P}^{\prime}}  \leftarrow {\mathbf{\hat{y}}_\mathcal{P'}^{ \top}}  \mathbf{y}$ \Comment{confidence scores}
\State $c \leftarrow \max(c_{\mathcal{P}^{'}}, (1-\lambda) c_{\mathcal{P}})$ \Comment{target confidence score}
\State $\tilde{\alpha} \; \leftarrow \; \dfrac{c - c^{}_{\mathcal{P}^{\prime}}}{c^{}_{\mathcal{P}} - c^{}_{\mathcal{P}^{\prime}}}$ \Comment{approximate $\alpha^\ast$ by Eq.~\eqref{eq:approx}}
\State $\mathcal{P}^{\ast} \; \leftarrow \; \tilde{\alpha} \mathcal{P} + (1-\tilde{\alpha}) \mathcal{P}^{\prime}$ \Comment{interpolate $\mathcal{P}$ and $\mathcal{P}^{\prime}$}
\end{algorithmic} 
\end{algorithm}
\vspace{-4pt}

\noindent\textbf{AugTune.} 
We present AugTune described in Algorithm \ref{algorithm_augtune} to control the strength of augmentation.
AugTune adjusts the strength of data augmentation by mixing the augmentation proposal $\mathcal{P'}$ from PointWOLF and the original sample $\mathcal{P}$.
Given a classifier $f(\cdot; \mathbf{w})$ and a sample $\mathcal{P}$, let $\mathbf{\hat{y}}_{\mathcal{P}}$ and $c_\mathcal{P}$ denote 
its prediction and confidence score, \ie, $\mathbf{\hat{y}}^{}_{\mathcal{P}}=f(\mathcal{P},\mathbf{w})$ and $c_\mathcal{P} = \mathbf{\hat{y}}^{\top}_{\mathcal{P}} \mathbf{y}$, where $\mathbf{y}$ is the ground truth label represented in one-hot encoding.
$\mathbf{\hat{y}}_{\mathcal{P}'}$ and $c_{\mathcal{P}'}$ are similarly defined for $\mathcal{P'}$.
Note that all the confidence scores $c_{\mathcal{P}}$, $c_{\mathcal{P}'}$ are obtained on the fly while training the model, \ie, an extra pretrained classifier is not required.
To adjust the strength of augmentation, given a \textit{difficulty coefficient} $\lambda \in (0, 1]$,
AugTune first computes a target confidence score $c$ for each sample by 
\begin{equation}
c = \max(c_{\mathcal{P'}}, (1-\lambda) c_{\mathcal{P}}).
\end{equation}
Assuming the augmented $\mathcal{P}'$ is difficult than the original $\mathcal{P}$, \ie, $c_\mathcal{P'}< c_{\mathcal{P}}$, as $\lambda$ gets close to 0, it implies that AugTune generates samples similar to the original sample $\Pc$.
Conversely, when $\lambda=1$, $c = c_{\mathcal{P'}}$, AugTune uses the augmentation proposal $\mathcal{P'}$ without any adjustment. 
To generate an augmented sample with the target confidence score, we use the linear interpolation of two samples $\mathcal{P}$ and $\mathcal{P'}$. 
Then, the problem is reduced to finding $\alpha^*$ defined by
\begin{equation}
\alpha^* = \argminU_{\alpha} \parallel c - f(\alpha \mathcal{P} + (1-\alpha) \mathcal{P}^{\prime})\parallel ^2.
\label{eq:optimal}
\end{equation}
However, solving \eqref{eq:optimal} directly by optimization algorithms or grid search is still computationally expensive. 
Thus, we approximate $\alpha^\ast$ by $\tilde{\alpha}=\frac{ c- c_{\mathcal{P}^{'}}}{c_{\mathcal{P}^{}}-c_{\mathcal{P}^{'}}}$ which is the solution to

    \begin{equation}
    \alpha c_\mathcal{P} + (1-\alpha) c_{\mathcal{P}^{\prime}} = c.
    \label{eq:approx}
    \end{equation}
Our experiments show this approximation does not cause degradation in the target tasks (see the supplement).
The final augmented sample $\mathcal{P}^\ast$ is a convex combination of $\mathcal{P}$ and $\mathcal{P}^\prime$  with $\tilde{\alpha}$, \ie, $\mathcal{P}^\ast = \tilde{\alpha} \mathcal{P} + (1-\tilde{\alpha}) \mathcal{P}^{\prime}$, then the model parameter $\mathbf{w}$ is updated as  $\mathbf{w} \; \leftarrow \; \mathbf{w} - \gamma\nabla_\mathbf{w}\mathcal{L}(f(\mathcal{P}^{\ast},\mathbf{w}), \mathbf{y})$, where $\gamma$ is a learning rate.
Note that since the correspondence between $\mathcal{P}$ and $\mathcal{P}'$  are known by construction, the interpolation of two point clouds can be obtained by a simple point-wise interpolation given as $\pb^\ast = \tilde{\alpha} \pb + (1-\tilde{\alpha})\pb'$.
Moreover, AugTune works as a safeguard to preserve the shape identity for the final $\mathcal{P}^\ast$.
So, we rarely observed unrealistic augmented samples with reasonable hyperparameters (see the supplement for visualizations).

\noindent\emph{Remarks.} As we viewed our framework as the kernel regression on local transformations, AugTune is directly applicable to the transformation (parameter) space.
In other words, instead of the point-wise interpolation, we may simply interpolate the local transformation parameters: scaling matrix $\mathbf{S}^\prime_j = \alpha\mathbf{I} + (1-\alpha)\mathbf{S}_j$, rotation $\theta^\prime_j = (1-\alpha)\theta_j$, and translation $\mathbf{b}^\prime_j = (1-\alpha)\mathbf{b}_j$.
However, due to its slightly higher computational cost and inferior performance, we applied AugTune on the input data space, see Section \ref{sec:4.2}.

\begin{table*}[ht]
  \centering
  \caption{\textbf{Overall mean IoU ($mIoU$) on ShapeNetPart.}}
\resizebox{\textwidth}{!}
{

\label{Partseg} 
\setlength{\tabcolsep}{3pt}
\renewcommand{\arraystretch}{1.05}
\begin{tabular}{c|cccccccccccccccc|c}
    \toprule
    \multirow{2}*{Method}&\multirow{2}*{\shortstack{air \\  plane}}&\multirow{2}*{bag}&\multirow{2}{*}{cap}&\multirow{2}{*}{car} &\multirow{2}{*}{chair}&\multirow{2}{*}{\shortstack{ear\\phone}}&\multirow{2}{*}{guitar}&\multirow{2}{*}{knife}&\multirow{2}{*}{lamp}&\multirow{2}{*}{labtop}&\multirow{2}{*}{\shortstack{motor\\bike}}&\multirow{2}{*}{mug}&\multirow{2}{*}{pistol}&\multirow{2}{*}{rocket}&\multirow{2}{*}{\shortstack{skate\\board}}&\multirow{2}{*}{table}&   \multirow{2}*{mIoU}  \\
    &&&&&&&&&&&&&&&&&\\
    \midrule
    PointNet&81.8&\textbf{74.7}&\textbf{80.2}&71.9&89.6&71.5&90.3&84.9&79.5&95.2&\textbf{65.2}&91.1&\textbf{81.1}&55.1&72.8&82.2&83.5 \\
    +PointWOLF&\textbf{82.5}&73.3&78.8&\textbf{73.2}&89.6&\textbf{72.2}&\textbf{91.2}&\textbf{86.2}&\textbf{79.7}&95.2&64.6&\textbf{92.5}&80.2&\textbf{56.6}&\textbf{73.1}&82.2&\textbf{83.8}\\
    
    \midrule
    PointNet++&81.9&83.4&86.4&\textbf{78.6}&90.5&64.7&\textbf{91.4}&83.1&83.4&95.1&\textbf{69.6}&\textbf{94.7}&\textbf{82.8}&56.9&\textbf{76.0}&\textbf{82.3}&84.8\\
    +PointWOLF&\textbf{82.0}&\textbf{83.9}&\textbf{87.3}&77.6&\textbf{90.6}&\textbf{78.4}&91.1&\textbf{87.6}&\textbf{84.7}&\textbf{95.2}&62.0&94.5&81.3&\textbf{62.5}&75.7&83.2&\textbf{85.2}\\
    \midrule
    DGCNN&82.2&\textbf{75.1}&81.3&\textbf{78.2}&90.6&73.6&90.8&87.8&84.4&95.6&\textbf{57.8}&92.8&\textbf{80.6}&51.5&73.9&82.8&84.8\\
    +PointWOLF&\textbf{82.9}&73.3&\textbf{83.5}&76.7&\textbf{90.8}&\textbf{76.7}&\textbf{91.4}&\textbf{89.2}&\textbf{85.2}&\textbf{95.8}&53.7&\textbf{94.0}&80.1&\textbf{54.9}&\textbf{74.3}&\textbf{83.4}&\textbf{85.2}\\
    \bottomrule
  \end{tabular}}
\end{table*}

\vspace{-5pt}
\section{Experiments}
In this section, we demonstrate the effectiveness of our PointWOLF on both synthetic and real-world datasets. 
We begin by describing the datasets, baselines, and experimental setup. 
Then, we evaluate our framework for shape classification and part segmentation (Section \ref{sec:4.1}), followed by ablation studies and analyses (Section \ref{sec:4.2}).
We conduct the experiments to show whether our method improves the robustness of models against both local and global corruptions leveraging diverse locally-augmented samples (Section \ref{sec:4.3}).
Lastly, we provide qualitative analysis of the augmented samples by PointWOLF (Section \ref{sec:4.4}).

\noindent\textbf{Datasets.}
We use both synthetic and real-world datasets for shape classification  to evaluate our framework. ModelNet40 (\textbf{MN40})~\cite{modelnet} is a widely used synthetic benchmark dataset containing 9,840 CAD models in the training set and 2,468 CAD models in the validation set with total 40 classes of common object categories. 
As in~\cite{pointmixup}, we also use the reduced version of MN40 (\textbf{ReducedMN40}) to simulate data scarcity.
ScanObjectNN (SONN) ~\cite{scanobjectnn} is a recent point cloud object dataset constructed from the real-world indoor datasets such as SceneNN~\cite{SceneNN} and ScanNet~\cite{scannet}. 
We use the following versions of SONN: (1) \textbf{OBJ\_ONLY} which has 2,309 and 581 scanned objects for the training and validation sets respectively and (2) \textbf{PB\_T50\_RS} which is a perturbed version with 11,416 and 2,882 scanned objects for the training and validation sets respectively. Both have 15 classes.
For part segmentation we adopt \textbf{ShapeNetPart}~\cite{shapenet}  which is a synthetic dataset contains 14,007 and 2,874 samples for training and validation sets. 
ShapeNetPart consists of 16 classes with 50 part labels. Each class has 2 to 6 parts.  

\noindent\textbf{Implementation Details.} 
All models and experiments are implemented in PyTorch. 
For PointNet and PointNet++, the PyTorch implementation by~\cite{pnpn++} was used with a minimum modification. 
With DGCNN, we use the official PyTorch code by the authors.
We train each model with a batch size of 32 for 250 epochs.
Note that for maximal fairness and consistency, we reproduced the numbers for every baselines except for PointMixup~\cite{pointmixup} and followed the evaluation protocol of \cite{pointmixup} for every case.
For our framework, the augmentation strength of PointWOLF was controlled by AugTune.
Indeed in our framework, the \textit{difficulty coefficient} $\lambda$ is the only hyperparameter to tune. We used $\lambda=0.1$ for synthetic datasets and $\lambda=0.3$ for all real-world datasets. For more details, see the supplement.

\noindent\textbf{Baselines.} 
We compare our framework (PointWOLF with AugTune) with the following data augmentation methods: \textbf{(1)} A conventional DA (\textbf{CDA}) that performs the global similarity transformation (\eg, rotation along the up-axis, scaling, and translation) with point-wise jittering as~\cite{pointnet2}. 
\textbf{(2)} \textbf{PointAugment} \cite{pointaugment} performs shape-wise transformation and point-wise displacement by learning an augmentor network. 
For datasets on which the models have not been evaluated in the literature, we use the authors' official implementation of \cite{pointaugment}.
\textbf{(3)} \textbf{PointMixup} \cite{pointmixup} uses the interpolated sample between two point clouds.

\subsection{Shape Classification and Part Segmentation}

\label{sec:4.1}
We evaluate our methods on shape classification using a synthetic dataset (MN40) and a real-world dataset (SONN). 
Also we conduct experiments on a synthetic dataset (ShapeNetPart) for part segmentation.

\noindent\textbf{Shape Classification.} 
Table \ref{MN40} shows that our PointWOLF achieves consistent improvements in overall accuracy on \textit{both MN40 and ReducedMN40 with all three models} compared to other augmentation methods (CDA, PointAugment, and PointMixup).
Observe that MN40 and ReducedMN40 are pre-aligned synthetic datasets and interestingly CDA without rotation denoted by CDA (w/o R) outperforms CDA. 
Despite the saturated datasets, PointWOLF improves overall accuracy by 1.6 \% compared to the best performing baseline on ReducedMN40 with PointNet.

Next, Table~\ref{SONN} shows the experimental results on SONN that is a more challenging
and diverse real-world dataset. 
As expected, diverse and realistic augmented samples from PointWOLF significantly improve the performance on \textit{both OBJ\_ONLY and PB\_T50\_RS with all three models}.
Specifically, on PB\_T50\_RS with PointNet++, 
the performance gains are 4.7\%, 6.2\%, and 3.5\% compared to  CDA, PointAugment, and PointMixup, respectively.
Our PointWOLF benefits the models more on the challenging cases with real-world data.

\noindent\textbf{Part Segmentation.} 
Given a point cloud $\mathcal{P}$, part segmentation is 
a point-wise classification where a model predicts a label for each point $\pb_i$.
In part segmentation, to derive the mixing ratio $\tilde{\alpha}$ in \eqref{eq:approx} at the object-level, we simply used the average of the pixel-wise confidence scores for our AugTune, \ie, $c_{\Pc} = \sum_i c_{\pb_i} / |\Pc|$. 
Our experiments in Table~\ref{Partseg} show that on ShapeNetPart~\cite{shapenet},
PointWOLF consistently improves mean IoU (mIoU) over baselines (0.3\% over PointNet, 0.4\% over PointNet++ and DGCNN), demonstrating the applicability of PointWOLF to point-wise tasks.
\begin{table}[t] 
  \centering
  \caption{\textbf{PointWOLF Ablation.} R: rotate, S: scale, T: translate.}
  \label{ablation3}
\setlength{\tabcolsep}{6pt}
\renewcommand{\arraystretch}{1.05}
  \begin{tabular}{c|ccc|c}
    \toprule
    \multicolumn{1}{c}{
    Local Transformation} & R & S & T & Accuracy\\
    \midrule
    None & & & & 86.6 \\
    +R &\checkmark & & &  88.1 \\
    +S & & \checkmark & & 88.6\\
    +T & & & \checkmark & 89.5\\
    +R, S, T&\checkmark&\checkmark &\checkmark& \textbf{89.7}\\
    \bottomrule
  \end{tabular}
\vspace{-5pt}
\end{table}

\vspace{-14pt}
\subsection{Analyses on PointWOLF and AugTune}
\label{sec:4.2}
We conduct ablation studies and analyses on SONN (OBJ\_ONLY) dataset with PointNet++ to analyze the significance of each component of PointWOLF and AugTune. 

\noindent\textbf{Local Transformation Ablations.} 
Table~\ref{ablation3} reports the ablations on three types of local transformations in PointWOLF: rotation (R), scaling (S), and translation (T).
PointWOLF with no local transformations is equivalent to PointNet++ \cite{pointnet2} with CDA. 
All three types of local transformations contribute to the accuracy gain. 
The best performance is obtained by +RST which utilizes all three local transformations, providing 3.5\% improvement over the baseline with no local transformations denoted by `None'.

\noindent\textbf{AugTune Ablations.} 
We evaluate how effectively AugTune controls the augmentation strengths given suboptimal augmentation ranges.
We set the augmentation ranges $S$ = ($\rho_r$=$15^{\circ}$, $\rho_s$=$2$, $\rho_t$=$1$)
and use the multiples of the augmentation ranges: $kS$=($k\rho_r$, $k\rho_s$, $k\rho_t$). 
Table~\ref{ablation2} shows that PointWOLF \textbf{w/ AugTune} outperforms PointWOLF \textbf{w/o AugTune} by 0.4 \% $\sim$ 1.9 \%. 
Our AugTune simplifies and accelerates the augmentation strength tuning with one \textit{difficulty coefficient} $\lambda$. 
Our AugTune also benefits other augmentation methods, e.g., CDA, (see the supplement).

\begin{table}[t] 
  \centering
  \caption{\textbf{Search Space Robustness Comparison.}}
  \label{ablation2}
\setlength{\tabcolsep}{7pt}
\renewcommand{\arraystretch}{1.05}
  \begin{tabular}{c|cc}
    \toprule
\multirow{1}{*}{\begin{tabular}[c]{@{}c@{}}Search Space\end{tabular}} & \multirow{1}{*}{w/o AugTune} & \multicolumn{1}{c}{w/ AugTune}               \\ 
     \midrule
     $S$ &88.8&\textbf{89.2}\\
    $2S$ &87.6&\textbf{88.6}\\
    $3S$ &86.1&\textbf{88.0}\\
    \bottomrule
  \end{tabular}
\end{table} 

\begin{table}[t] \small
  \centering
\renewcommand{\arraystretch}{1}
  \caption{\textbf{Interpolation space for AugTune.}}
  \label{interpolation_space}
    \setlength{\tabcolsep}{7pt}
    \renewcommand{\arraystretch}{1.05}

  \begin{tabular}{ccc}
    \toprule
    Space & Accuracy & Complexity\\
    \midrule
    Transformation Space&88.1& $O(MN)$\\
    Input Data Space&89.7&$O(N)$\\
    \bottomrule
\end{tabular}
\end{table}

\noindent\textbf{Interpolation Space for AugTune.}
Two interpolation spaces can be considered for AugTune: the input data space and the transformation (parameter) space.
Although directly tuning the transformation parameters seems natural, we have experimentally shown that AugTune in the input data space is a sensible choice.
Table~\ref{interpolation_space} shows the superiority of AugTune in the input data space regarding both performance and computational efficiency.
For $N$ points and $M$ anchor points, AugTune in the transformation (parameter) space requires computing a new transformation for each point and each anchor point in $O(MN)$. 
Contrarily, AugTune in the input data space simply interpolates the points (\ie, $\alpha \pb + (1-\alpha)\pb'$ for each $\pb$) in $O(N)$.

\subsection{Robustness to Corruption}
\label{sec:4.3}
\begin{figure}[t] 
\centering
\subfigure[LocalDrop]{
    \label{subfig:3a}
    \includegraphics[trim=30 0 60 0,clip,height=2.15cm]{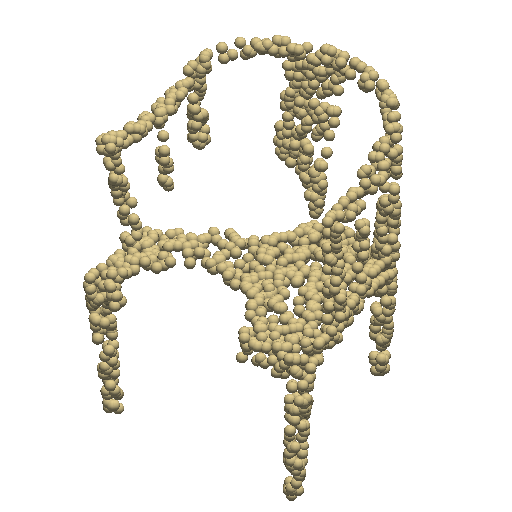}%
}
\subfigure[LocalAdd]{
    \label{subfig:3b}
    \includegraphics[trim=60 30 70 0,clip,height=2.5cm]{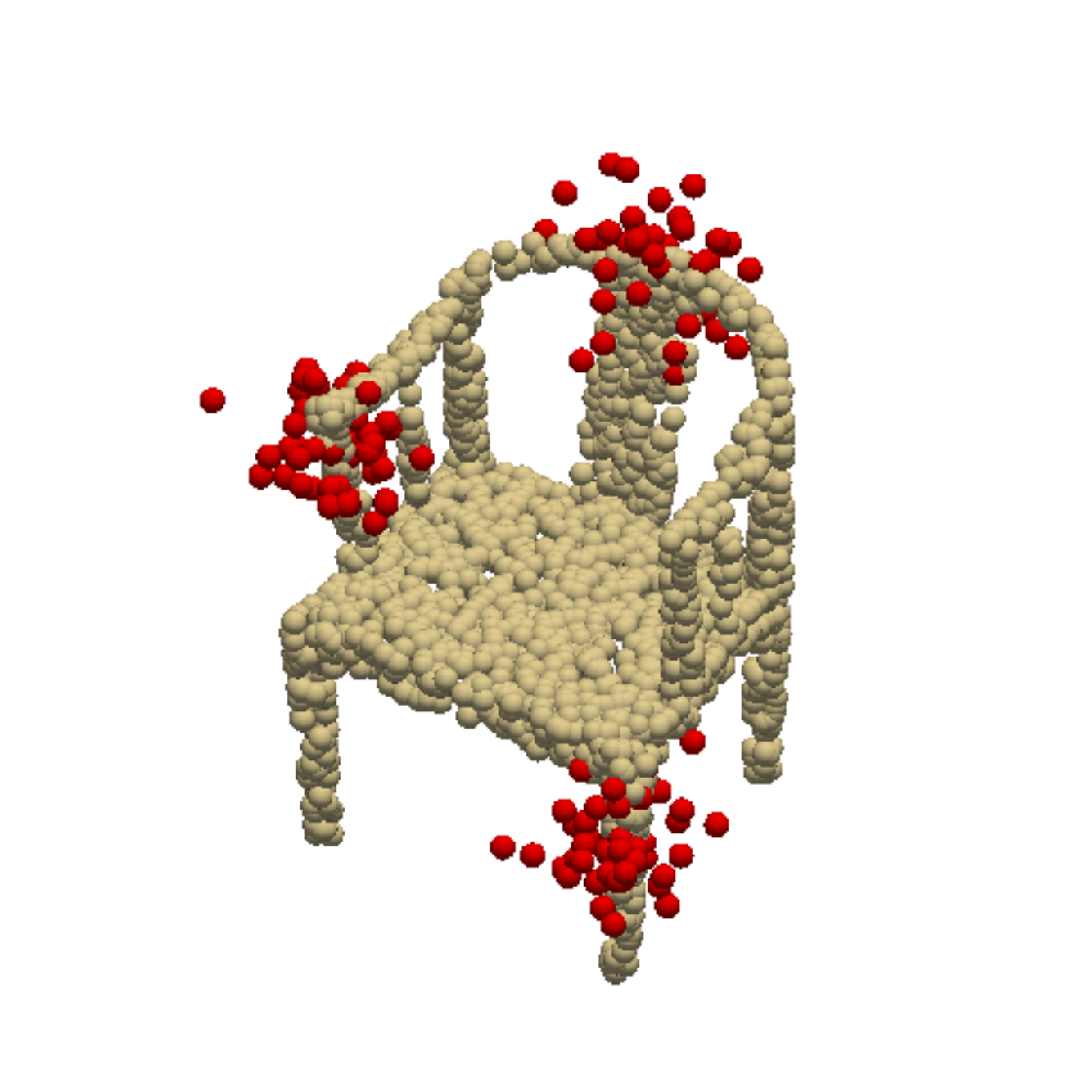}%
}
\subfigure[Dropout]{
    \label{subfig:3c}
    \includegraphics[trim=50 30 110 0,clip,height=2.15cm]{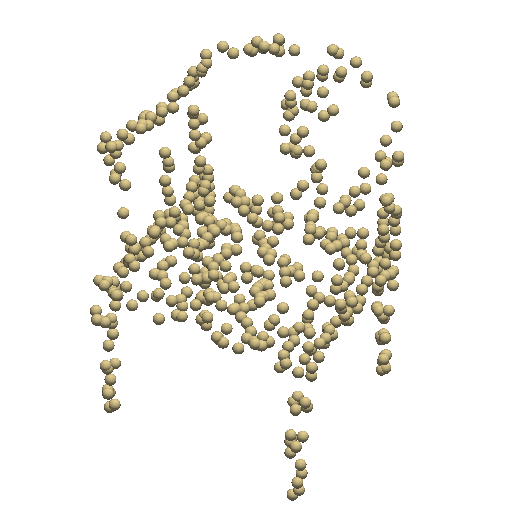}%
}
\subfigure[Noise]{
    \label{subfig:3d}
    \includegraphics[trim=50 30 120 0,clip,height=2.35cm]{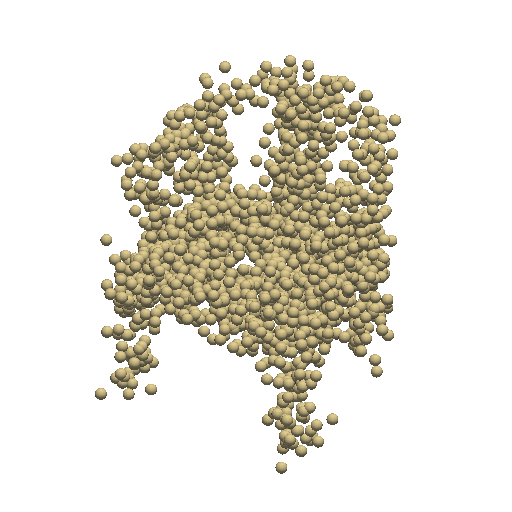}
}

\caption{\textbf{Illustration of Local and Global Corruption.} (a) and (b) are local corruptions while (c) and (d) are global corruptions.}
\label{fig:fig3}
\end{figure}

Additional studies demonstrate our PointWOLF improves the robustness of models against various corruptions as shown in Figure~\ref{fig:fig3}.
First, we consider two \textit{local} corruptions: (1) \textbf{LocalDrop} drops $\mathcal{C}$ local clusters and (2) \textbf{LocalAdd} adds $\mathcal{C}$ local clusters where a cluster consists of $K$ nearest points from a randomly selected cluster center point.
We used $K=50$ in both cases.
Second, to examine the general robustness to global corruption, we perform random point-wise (3) \textbf{Dropout} with a dropout rate $r \in \{0.25, 0.5, 0.75\}$ and (4) \textbf{Noise} perturbation by offsets drawn from a Gaussian distribution with standard deviation $\sigma \in \{0.01,0.03,0.05\}$.

We trained PointNet++ with CDA (baseline) and PointWOLF and evaluated them on corrupted samples by the local and global corruptions above.
Experimental results on MN40 in Table~\ref{distortion} show that compared to CDA, PointWOLF consistently and significantly improves the robustness against various corruptions.
Importantly, the gain over the baseline significantly increases as the amount of corruptions increases: 7.2\% for LocalDrop ($\mathcal{C}=7$), 13.1\% for LocalAdd ($\mathcal{C}=7$), 31.3\% for Dropout ($r=0.75$), and 22.2\% for Noise ($\sigma=0.05$). 
We believe that the diverse samples augmented by locally weighted transformations in PointWOLF help models to learn more robust features against both `local' and `global' corruptions.

\begin{table}[t] 
  \centering
\setlength{\tabcolsep}{7pt}
\renewcommand{\arraystretch}{1.05}
  \caption{\textbf{Robustness to Corruption.}}
  \label{distortion}
  \begin{tabular}{ll|cl}
    \toprule
    \multicolumn{2}{c}{Corruption} & \multicolumn{1}{c}{CDA} & \multicolumn{1}{c}{PointWOLF} \\
    \midrule
    \multirow{3}*{LocalDrop} & $\mathcal{C}$=3 & 67.0 & 68.8 (1.8$\uparrow$)\\
    &$\mathcal{C}$=5& 63.2 & 66.5 (3.3$\uparrow$)\\
    &$\mathcal{C}$=7& 52.8 & 60.0 (7.2$\uparrow$)\\
    \midrule
    \multirow{3}*{LocalAdd} & $\mathcal{C}$=3 & 73.9 & 77.2 (3.3$\uparrow$)\\
    &$\mathcal{C}$=5& 63.5 & 69.4 (5.9$\uparrow$)\\
    &$\mathcal{C}$=7& 51.5 & 64.6 (13.1$\uparrow$)\\
    \midrule
    \multirow{3}*{Dropout} & $r$=0.25 & 91.2 & 92.2 (1.0$\uparrow$)\\
    &$r$=0.5 & 84.0 & 90.4 (6.4$\uparrow$)\\
    &$r$=0.75 & 29.5 & 60.8 (31.3$\uparrow$)\\
    \midrule
    \multirow{3}*{Noise} & $\sigma$=0.01 & 91.5 & 93.0 (1.5$\uparrow$)\\
    &$\sigma$=0.03 & 78.8 & 87.6 (8.8$\uparrow$)\\
    &$\sigma$=0.05 & 22.9 & 45.1 (22.2$\uparrow$)\\
    \bottomrule
  \end{tabular}
\end{table}

\subsection{Qualitative Analysis}
\label{sec:4.4}
\begin{figure}[t] 
\centering
\subfigure[Torsion]{
    \label{subfig:4a}
    \includegraphics[width=3.6cm, height=3.6cm]{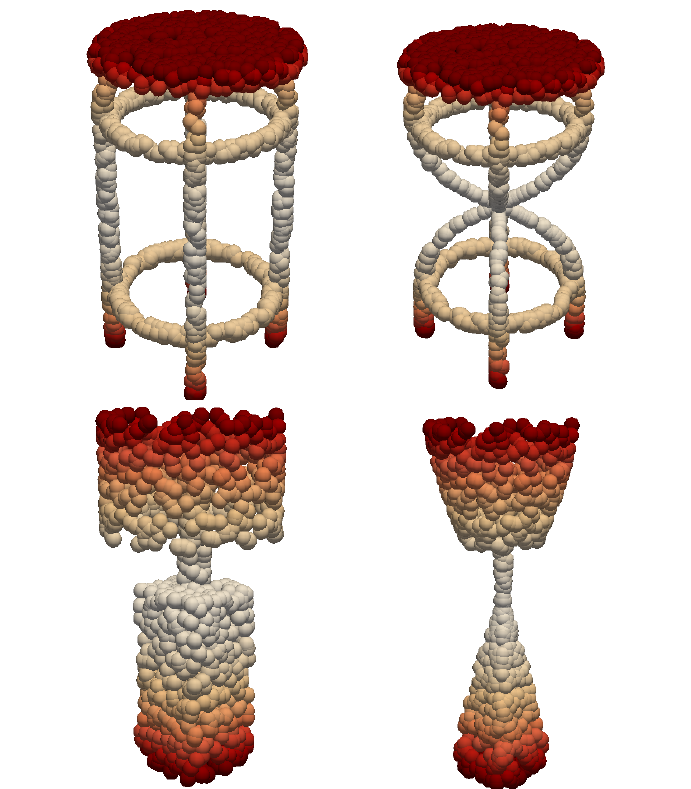}%
}
\subfigure[Shearing]{
    \label{subfig:4b}
    \includegraphics[width=3.6cm, height=3.6cm]{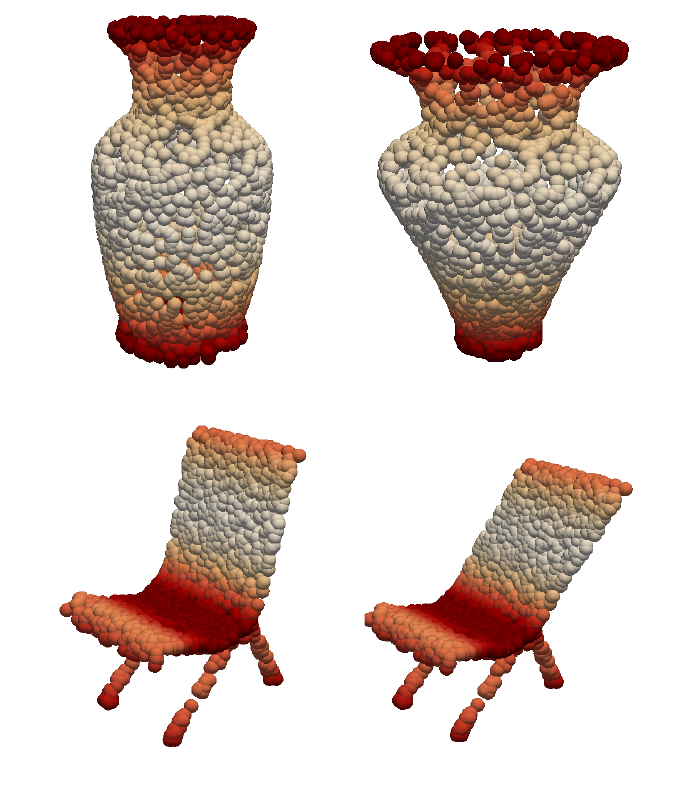}
}

\subfigure[Partial Scaling]{
    \label{subfig:4c}
    \includegraphics[trim=0 0 0 0,clip,width=3.6cm, height=3.6cm]{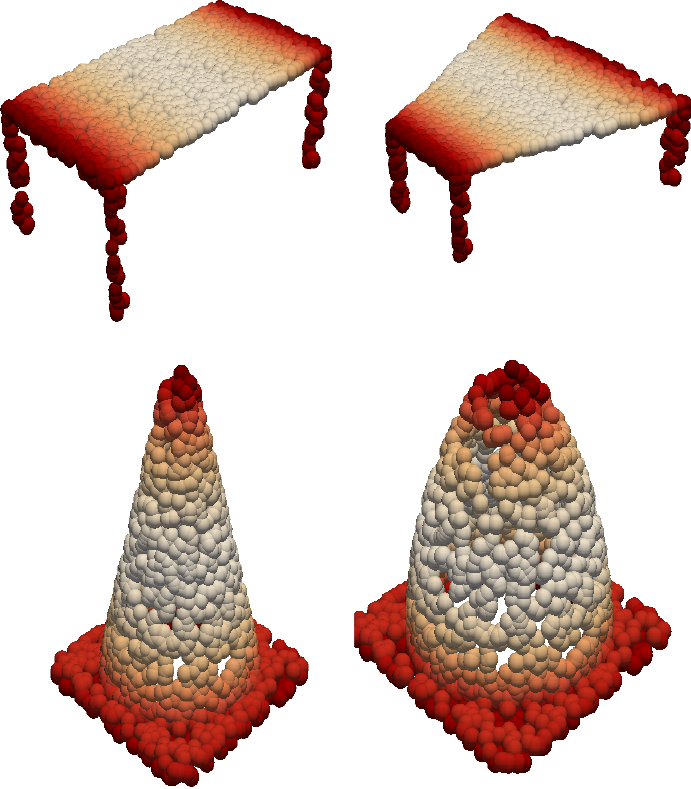}%
}
\subfigure[Combination]{
    \label{subfig:4d}
    \includegraphics[trim=0 0 0 0
    ,clip,width=3.6cm, height=3.6cm]{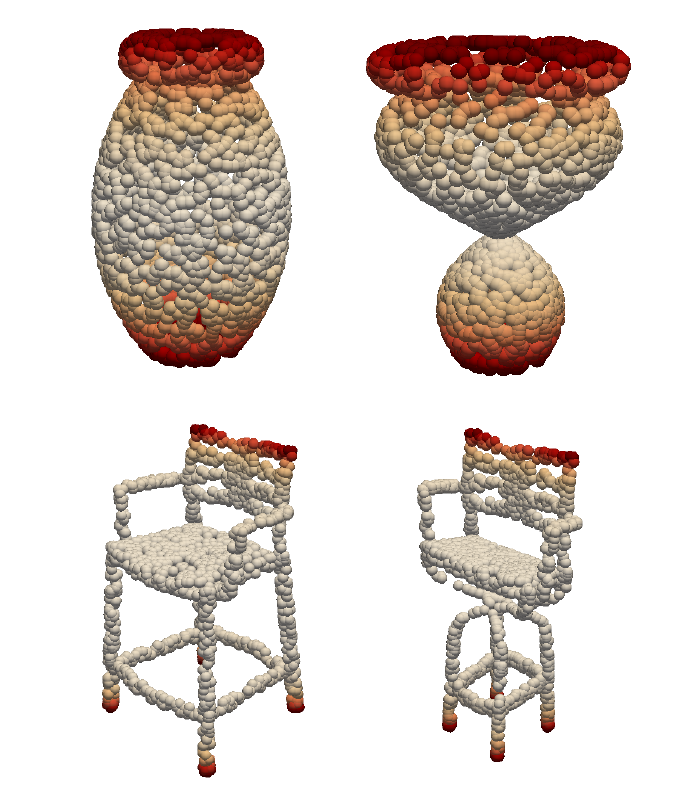}
}
\caption{\textbf{Advanced Deformations by PointWOLF.} 
In each transformations, the locally transformed samples (right) are generated from original samples (left). 
}
\vspace{-1pt}
\label{fig:fig4}
\end{figure}

Although PointWOLF essentially makes use of simple transformations such as rotation, scaling, and translation, we interestingly find that PointWOLF often mimics highly advanced yet realistic global deformations like torsion and shearing which \textit{cannot} trivially be applied to point clouds.
We achieve this by (1) projecting the transformations to random subsets of the axes and (2) allowing AugTune to identify ``beneficial'' cases which interestingly turn out to be a set of realistic deformations.
Figure~\ref{fig:fig4} displays several such examples.
For instance, when two anchor points are located at the top and bottom of the stool in Figure~\ref{subfig:4a}, a \textit{torsion} occurs when it rotates only along the up-axis while preserving the near-anchor shapes of bright regions.

Similarly, a combination of local scaling and translation produce \textit{shearing} or \textit{partial scaling}.
In fact, many advanced deformations that naturally preserve the shape identity are commonly defined by combinations of simpler transformations.
In this sense, PointWOLF can adaptively allow a set of local transformations that often mimic advanced deformations.
Importantly, seeing how these visually explainable augmentations from local transformations also bring empirical benefits, understanding and exploiting local structures are crucial for successful DA on point cloud.
\section{Conclusion}
We propose a novel point cloud augmentation method, PointWOLF, which augments point clouds by weighted local transformations.
Our method generates diverse and realistic augmented samples with smoothly varying deformations formulated as a kernel regression and brings significant improvements on point cloud tasks across several datasets.
Moreover, to find an optimal augmentation in an expansive search space, our AugTune adaptively controls the strength of augmentation during training with a single hyperparameter. 
Our findings show that the augmentations we produce are not only visually realistic but also beneficial to the models, further validating the importance of understanding the local structure of point clouds.

\vspace{-10pt}
\paragraph*{Acknowledgments.}
This work was supported by ICT Creative Consilience program(IITP-2021-2020-0-01819) supervised by the IITP, Research on CPU vulnerability detection and validation (No. 2019-0-00533), the National Research Council of Science \& Technology (NST) grant by the Korea government (MSIT) (No. CAP-18-03-ETRI), and Samsung Electronics.

{\small
\bibliographystyle{unsrt}
\bibliography{egbib}
}
\end{document}